\documentclass{ecai}
\usepackage{times}
\usepackage{graphicx}
\usepackage{latexsym}
\usepackage{xcolor}

\begin{document}

\title{Unsupervised Vehicle Counting via Multiple Camera Domain Adaptation \footnote{Copyright © 2020 for this paper by its authors. Use permitted under
  Creative Commons License Attribution 4.0 International (CC BY 4.0).}}

\author{Luca Ciampi\institute{Institute of Information Science and Technologies (ISTI), Italian National Research Council (CNR),Italy, Pisa. Email: luca.ciampi@isti.cnr.it} \and
Carlos Santiago\institute{Instituto Superior T\'ecnico (LARSyS/IST), Portugal, Lisbon.} \and
Joao Paulo Costeira$^2$ \and \\  
Claudio Gennaro$^1$\and 
Giuseppe Amato$^1$}

\maketitle
\bibliographystyle{ecai}

\begin{abstract}
Monitoring vehicle flows in cities is crucial to improve the urban environment and quality of life of citizens. Images are the best sensing modality to perceive and assess the flow of vehicles in large areas. Current technologies for vehicle counting in images hinge on large quantities of annotated data, preventing their scalability to city-scale as new cameras are added to the system. This is a recurrent problem when dealing with physical systems and a key research area in Machine Learning and AI. We propose and discuss a new methodology to design image-based vehicle density estimators with few labeled data via multiple camera domain adaptations.  
\end{abstract}

\section{INTRODUCTION}

Artificial Intelligence (AI) systems dedicated to the analysis and interaction with the physical world can significantly impact human life. These systems can process a massive amount of data and make/suggest decisions that help solve many real-world problems where humans are at the epicenter.

Crucial examples of human-centered artificial intelligence, whose aim is to create a better world by achieving common goals beneficial to our societies, are city mobility, pollution monitoring, or critical infrastructure management, where decision-makers require, for instance,  measurements about flows of bicycles, cars or people.
Like no other sensing mechanism, networks of city cameras can observe such large dimensions and simultaneously provide visual data to AI systems to extract relevant information from this deluge of data.

Different smart cameras across the city are subject to various visual conditions (luminance, position, context). This results in different performances from each of them and added difficulty in effectively scaling-up the learning task. In this paper, we address this issue proposing a methodology that performs unsupervised domain adaptation among different cameras to compute the number of vehicles in a city reliably. We focus on vehicle counting, but the approach is applicable to counting any other type of object.


\subsection{Counting as a supervised learning task}

\begin{figure}[t]
\centerline{\includegraphics[width=.4\textwidth]{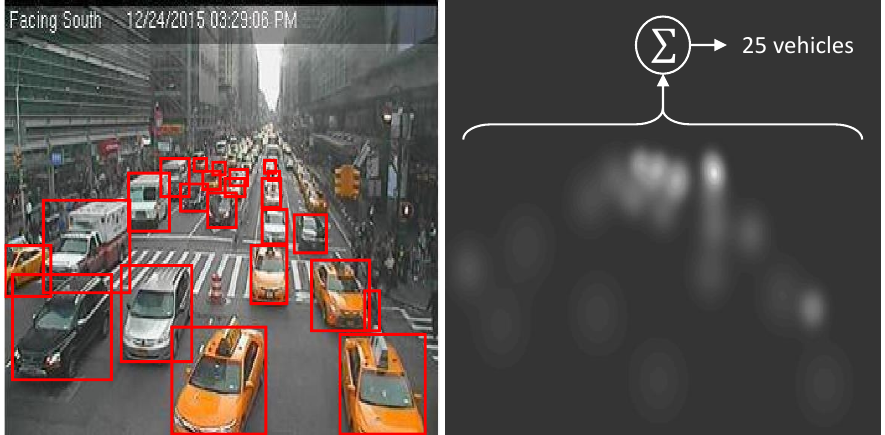}}
\caption{Example of an image with the bounding box annotations (left) and the corresponding density map that sums to the counting value (right).} 
\label{fig:density}
\end{figure}

The counting problem is the estimation of the number of objects instances in still images or video frames \cite{lempitsky2010learning}. Current systems address the counting problem as a supervised learning process. They fall in two main classes of methods: a) detection-based approaches (\cite{amato2019counting, ciampi2018counting, amato2018wireless}) that try to identify and localize single instances of objects in the image and b)density-based techniques that rely on regression techniques to estimate a density map from the image, and where the final count is given by summing all pixel values~\cite{lempitsky2010learning}. Figure \ref{fig:density} illustrates the mapping of such regression. Concerning vehicle counting in urban spaces, where images are of very low resolution, and most objects are partially occluded, density-based methods have a clear advantage on detection methods~\cite{zhang2016single, guerrero2015extremely, li2018csrnet, boominathan2016crowdnet}.  


Hinging on Convolution Neural Networks (CNN) to learn the regressor, this class of approaches has shown to be very effective, especially in single-camera scenarios. However, since they require pixel-level ground truth for supervised learning, they may not generalize well to unseen images, especially when there is a large  \textit{domain gap} between the training (\textit{source}) and the test (\textit{target}) sets, such as different camera perspectives, weather, or illumination. This gap severely hampers the application of counting methods to very large scale scenarios since annotating images for all the possible cases is unfeasible.

\subsection{Unsupervised domain adaptation}
This paper proposes to generalize the counting process through a new domain adaptation algorithm for density map estimation and counting. Specifically, we suppose to have an annotated training set for a \emph{source domain}, and we want to adapt the system to perform well in an unseen and unlabelled \emph{target domain}. For instance, the source domain consists of images taken from a set of cameras. In contrast, the target domain consists of pictures taken from different cameras, with different luminances, perspectives, and contexts. This class of algorithms is commonly referred to as \textit{Unsupervised Domain Adaptation}.

We conduct preliminary experiments using the \textit{WebCamT} dataset introduced in \cite{zhang2017understanding}. In particular, we consider a test set containing images from cameras with different perspectives from the training ones, showing that our unsupervised domain adaptation technique can mitigate the perspective domain gap. 

Traditional approaches of Unsupervised Domain Adaptation have been developed to address the problem of image classification, and they try to align features across the two domains (\cite{ganin2014unsupervised, tzeng2017adversarial}). However, as pointed out in \cite{zhang2017curriculum}, they do not perform well in other tasks, such as semantic segmentation.


\section{Proposed Method}

We propose an end-to-end CNN-based unsupervised domain adaptation algorithm for traffic density estimation and counting. Inspired by \cite{tsai2018learning}, we base our method on adversarial learning in the \textit{output} space (density maps), which contains rich information such as scene layout and context. In our approach, we rely on the adversarial learning scheme to make the predicted density distributions of the source and target domains consistent. 

\begin{figure*}
\centerline{\includegraphics[width=.59\textwidth]{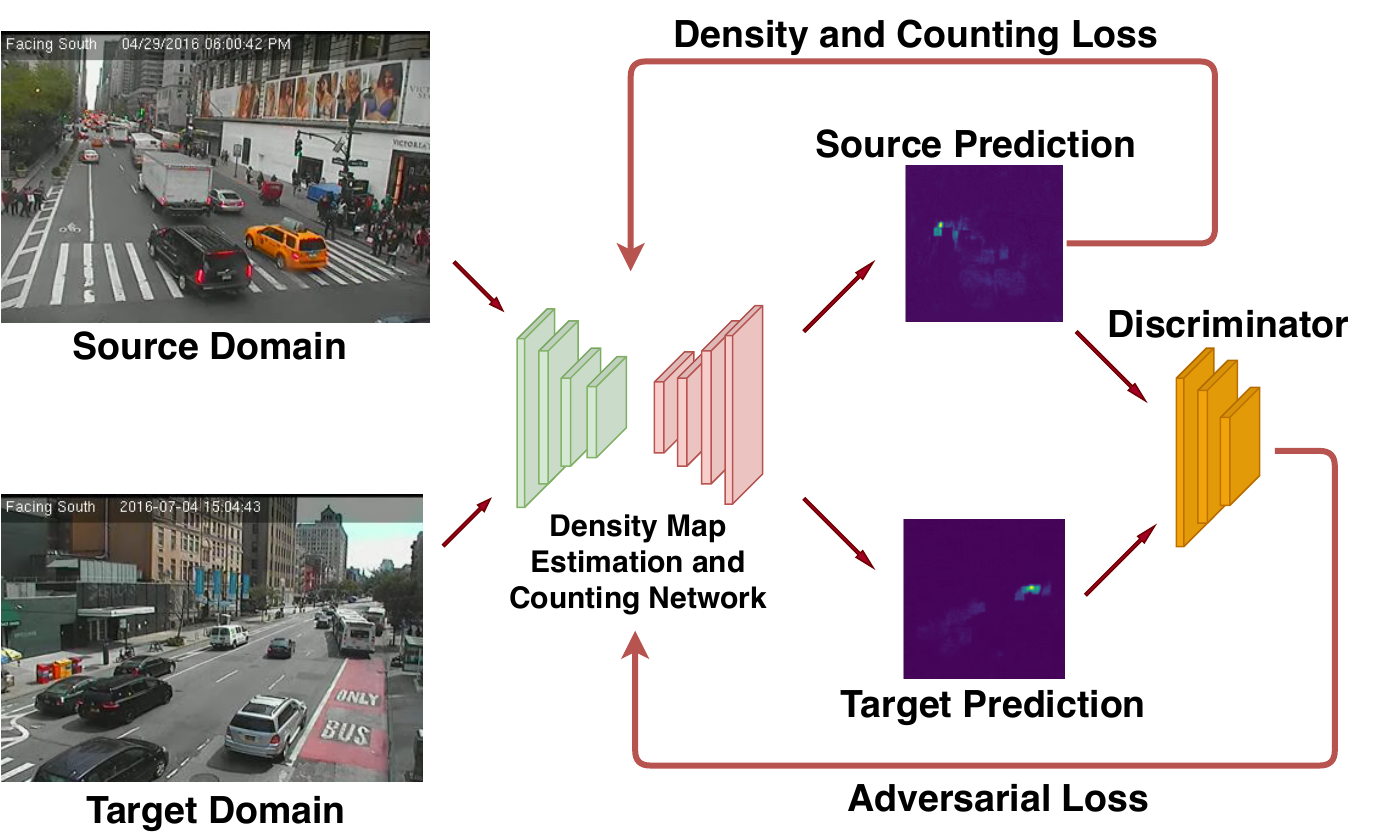}}
\caption{Algorithm overview. Given images having size \(C \times H \times W\) from source and target domains, we pass them through the density map estimation and counting network to obtain output predictions. For source predictions, a density and counting loss is computed based on the source ground truth. To make target predictions closer to the source ones, we employ a discriminator that aims to distinguish whether the input (i.e., the density map) belongs to the source or target domain. Then an adversarial loss is computed on the target prediction and is back-propagated to the density map estimation and counting network.}
\label{overview_approach}
\end{figure*}

The proposed framework, shown in Fig. \ref{overview_approach}, consists of two modules: 1) a CNN that predicts traffic density maps and estimates the number of vehicles occurring in the scene, and 2) a discriminator that distinguishes whether the density map (received by the density map estimator) is generated processing an image of the source domain or the target domain. In the training phase, the density map predictor learns to map images to densities, based on annotated data from the source domain. At the same time, it learns to fool the discriminator exploiting an adversarial loss, computed using the predicted density map of unlabeled images from the target domain. Consequently, the output space is forced to have similar distributions for both the source and target domains. In the inference phase, the discriminator is discarded, and only the density map predictor is used for the target images. A description of each module and their training is provided in the following subsections.

\subsection{Density Estimation Network}

We formulate the counting task as a density map estimation problem \cite{lempitsky2010learning}. The density (weight) of each pixel in the map depends on its proximity to a vehicle centroid and the size of the vehicle in the image, as shown in Fig. \ref{fig:density}, so that each vehicle contributes with a total value of 1 to the map. Therefore, it provides statistical information about the vehicles' location and allows the counting to be estimated by summing of all density values.


This task is performed by a CNN-based model, whose goal is to automatically determine the vehicle density map associated with a given input image.
Formally, the density map estimator, \(\Psi: \mathcal{R^{C \times W \times H}} \mapsto \mathcal{R^{W \times H}}\), transforms a \(\mathcal C\) channels \(\mathcal W \times \mathcal H\) input image, \(\mathcal{I}\), into a density map, \(D=\Psi(\mathcal{I})\in\mathcal{R^{W \times H}}\).



\subsection{Discriminator Network}
The discriminator network, denoted by \(\Theta\), also consists of a CNN model. It takes as input the density map, \(D\), estimated by the network \(\Psi\). Its output is a lower resolution probability map. Each pixel represents the probability that the corresponding area (from the input density map) comes from the source or the target domain. The goal of the discriminator is to learn to distinguish between density maps belonging to source or target domains. This, in turn, forces the density estimator to provide density maps with similar distributions in both domains, \textit{i.e.}, the density maps, \(D\), of the target domain have to look realistic, even if network \(\Psi\) was not trained with an annotated training set from that domain.

\subsection{Domain Adaptation Learning}

The proposed framework is trained based on an alternate optimization of density estimation network, \(\Psi\), and the discriminator network, \(\Theta\). Regarding the former, the training process relies on two components: 1) density estimation using pairs of images and ground truth density maps, which we assume are only available in the source domain; and 2) adversarial training, which aims to make the discriminator fail to distinguish between the source and target domains. As for the latter, images from both domains are used to train the discriminator on correctly classifying each pixel of the probability map as either source or target.

To implement the above training procedure, we introduce two loss functions: one is employed in the first step of the algorithm to train network \(\Psi\). The other is used in the second step to train the discriminator \(\Theta\). These loss functions are detailed next.

\textbf{Network \(\Psi\) Training.} We formulate the loss function for \(\Psi\) as the sum of two main components:

\begin{equation}
\mathcal{L}(\mathcal{I^\mathcal{S}}, \mathcal{I^\mathcal{T}}) = \mathcal{L}_{density}(\mathcal{I^\mathcal{S}}) + \lambda_{adv}\mathcal{L}_{adv}(\mathcal{I^\mathcal{T}}),
\label{global_loss_equation}
\end{equation}

\noindent where \(\mathcal{L}_{density}\) is a composite loss computed using ground truth annotations available in the source domain, while \(\mathcal{L}_{adv}\) is the adversarial loss that is responsible for making the distribution of the target and the source domain close each other. In particular, we define the density loss \(\mathcal{L}_{density}\) as:

\begin{equation}
\mathcal{L}_{density}(\mathcal{I^\mathcal{S}}) = \mathcal{L}_{density\_map}(\mathcal{I^\mathcal{S}}) + \mathcal{L}_{regression}(\mathcal{I^\mathcal{S}}),
\end{equation}

\noindent where \(\mathcal{L}_{density\_map}\) is the mean square error between the predicted and ground truth density maps, i.e. \(\mathcal{L}_{density\_map} = MSE(D^{\mathcal{S}}, D^{\mathcal{S\_GT}})\), while \(\mathcal{L}_{regression}\) is Euclidean loss between predicted and ground truth count.

To compute the adversarial loss \(\mathcal{L}_{adv}(\mathcal{I^\mathcal{T}})\), we first forward the images belonging to the target domain and we generate the predicted density maps \(D^{\mathcal{T}}\). Then, we compute

\begin{equation}
\mathcal{L}_{adv}(\mathcal{I^\mathcal{T}}) = - \sum_{h, w}\log (\Theta(D^{\mathcal{T}})).
\end{equation}

\noindent This loss forces the distribution of \(D^{\mathcal{T}}\) to be closer to \(D^{\mathcal{S}}\) by training \(\Psi\) to fool the discriminator, maximizing the probability of the target predicted density map to be considered as the source prediction.

\textbf{Discriminator \(\Theta\) Training.} Given the estimated density map \(D = \Psi(\mathcal{I})\in\mathcal{R^{W \times H}}\), we forward \(D\) to a fully-convolutional discriminator \(\Theta\) using a binary cross-entropy loss \(\mathcal{L}_{disc}\) for the two classes (i.e., source and target domains). We formulate the loss as:

\begin{equation}
\mathcal{L}_{disc}(D) = - \sum_{h, w}[(1-y)\log(\Theta(D)^{(h,w,0)}) + y\log(\Theta(D)^{(h,w,1)})],
\end{equation}

\noindent where \(y = 0\) if the sample is taken from the target domain, and \(y = 1\) if the sample is taken from the source domain.

\subsection{Implementation Details}
\textbf{Density Map Estimation and Counting Network.} We build our density map estimation network based on U-Net \cite{ronneberger2015u}. U-Net is a popular end-to-end encoder-decoder network for semantic segmentation first used for biomedical image segmentation. The encoder part consists of convolution blocks, followed by max-pooling blocks that downscale the feature representations at multiple levels. The decoder part of the network upsamples the features through upsampling layers followed by regular convolution operations. Furthermore, the upsampled features are concatenated with the same scale features from the encoder, containing more detailed spatial information and preventing the network from losing spatial awareness due to downsampling.

\noindent\textbf{Discriminator.} We use a Fully Convolutional Network similar to \cite{tsai2018learning, radford2015unsupervised}, composed of 5 convolution layers with kernel \(4\times4\) and stride of 2. The number of channels are \{64, 128, 256, 512, 1\}, respectively. Each convolution layer is followed by a leaky ReLU having a parameter equals to 0.2. 

\section{EXPERIMENTAL SETUP}
We conduct preliminary experiments using the \textit{WebCamT} dataset introduced in \cite{zhang2017understanding}. This dataset is a collection of traffic scenes recorded using city-cameras, and it is particularly challenging for analysis due to the low-resolution \((352\times240)\), high occlusion, and large perspective. We consider a total of about 42,000 images belonging to 10 different cameras and consequently having different perspectives. We employ the existing bounding box annotations of the dataset to generate ground truth density maps, one for each image. In particular, we consider one Gaussian Normal kernel for each vehicle present in the scene, having a value of \(\mu\) and \(\sigma\) equals to the center and proportional to the length of the bounding box surrounding the vehicle, respectively. 

Firstly, we show the domain gap that we want to face. We generate a first pair of training and validation subsets, picking images randomly from the whole dataset. Then, we create a second pair of training and validation subsets, this time picking images belonging to seven different cameras for the first and pictures belonging to the three remaining ones for the second (\textit{per-camera} splits of the whole dataset). We show the domain gap training our model without the discriminator on the training subsets and comparing the results obtained over the validation splits. 

Once we quantified and proved this domain gap, we try to mitigate it, conducting experiments on the per-camera splits using our solution, i.e., the network \(\Psi\) and the discriminator \(\Theta\) that acts on the output space. In particular, during the training, we also use the images belonging to the validation subset \textit{without} the labels to generate an adversarial loss aimed at making the source domain (i.e., the training subset) and the target domain (i.e., the validation subset) close each other.

We base the evaluation of the models on three metrics: (i) Mean Absolute Error (MAE) that measures the absolute count error of each image; (ii) Mean Squared Error (MSE) that penalizes large errors more heavily than small ones; (iii) Average Relative Error (ARE), which measures the absolute count error divided by the true count.

\section{RESULTS AND DISCUSSION}

\begin{figure}[t]
\centering
\begin{tabular}{cc}
\includegraphics[width=.21\textwidth]{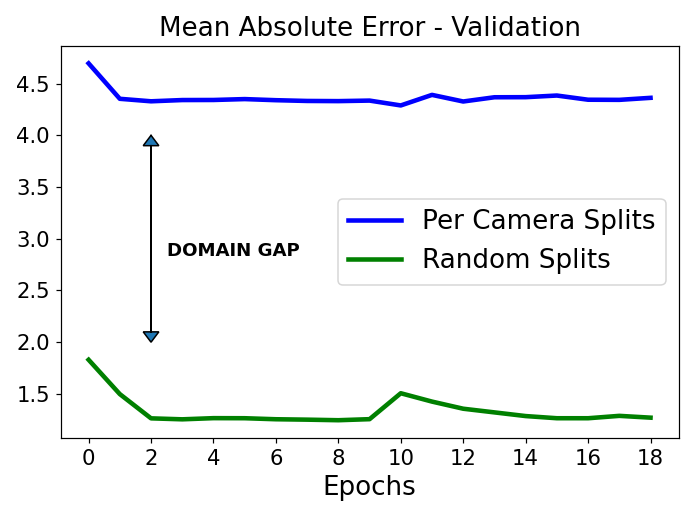} & 
\includegraphics[width=.21\textwidth]{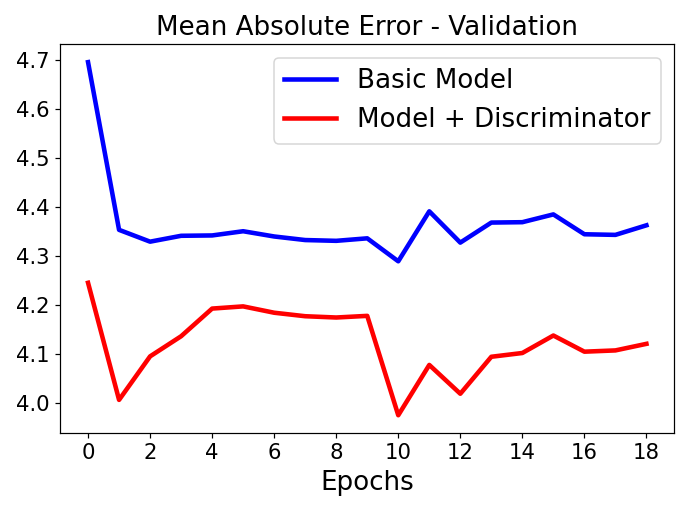}  \\
\includegraphics[width=.21\textwidth]{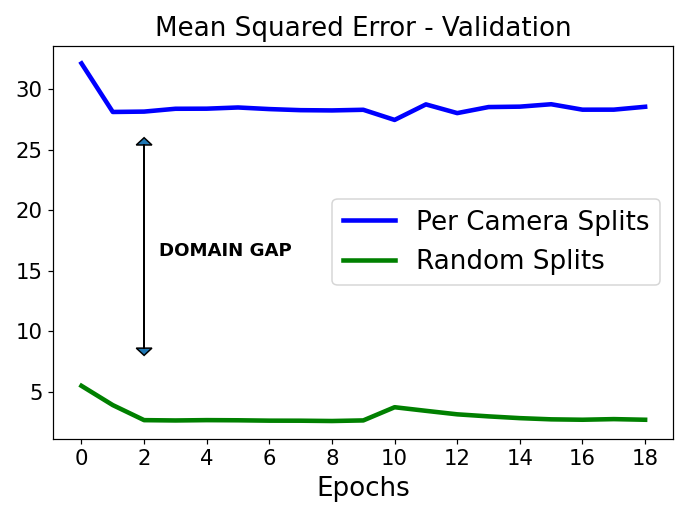} & 
\includegraphics[width=.21\textwidth]{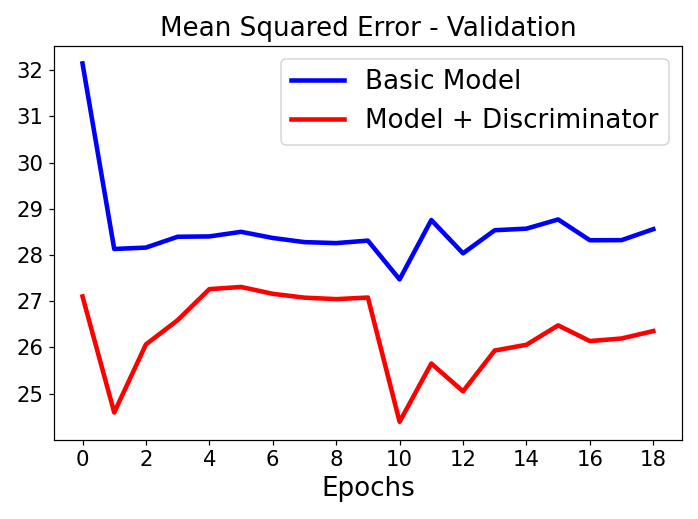} \\
\includegraphics[width=.21\textwidth]{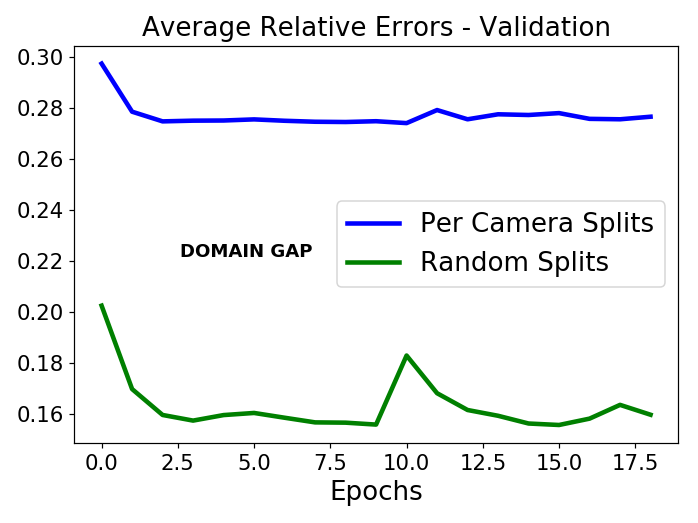} & 
\includegraphics[width=.21\textwidth]{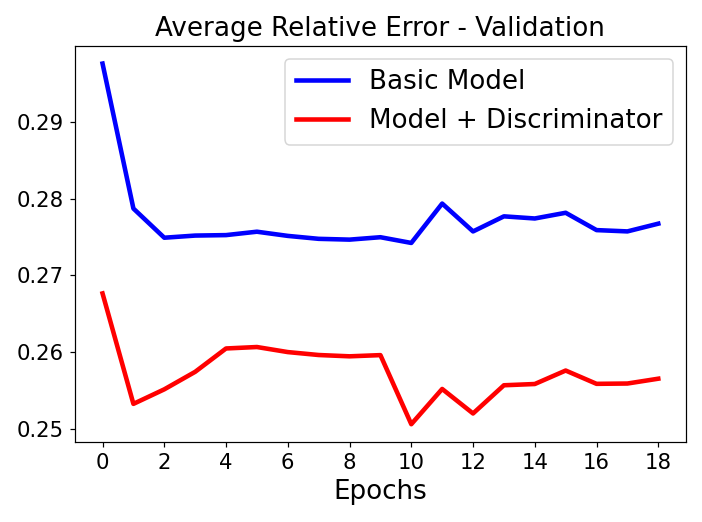}\\
   (a)  &  (b)
\end{tabular}
\caption{Performance during training: (a) Comparison between the random and the per-camera validation splits showing the domain gap; (b) comparison between the proposed approach with and without discriminator. Each row corresponds to a specific evaluation metric.} 
\label{plot_results}
\end{figure}

Figure \ref{plot_results} (a) shows the results for the two validation sets - the random one and the per-camera one, using the density estimation network without the discriminator trained over the two training subsets - the random one and the per-camera one, respectively. Each plot corresponds to one of the three metrics. As we can see, the domain gap is significant: even if all the subsets' images belong to the same dataset and are collected in the same city under similar conditions, small changes to the perspectives cause a remarkable loss in performance. In other words, the network cannot generalize well to views that have not been seen during the training.

When combining the density estimation network with the adversarial component, the performance of the system improves considerably. These results are shown in Figure \ref{plot_results} (b), where the improvements obtained using our model (red line) compared to the baseline model, without discriminator, is visible in all the three metrics. The discriminator mitigates the domain gap, and the network can generalize better over images having different perspectives from the ones employed during the training. The results are related to a specific value of \(\lambda\) that showed the most promising results.

\begin{figure}[t]
\centerline{\includegraphics[height=1.75in]{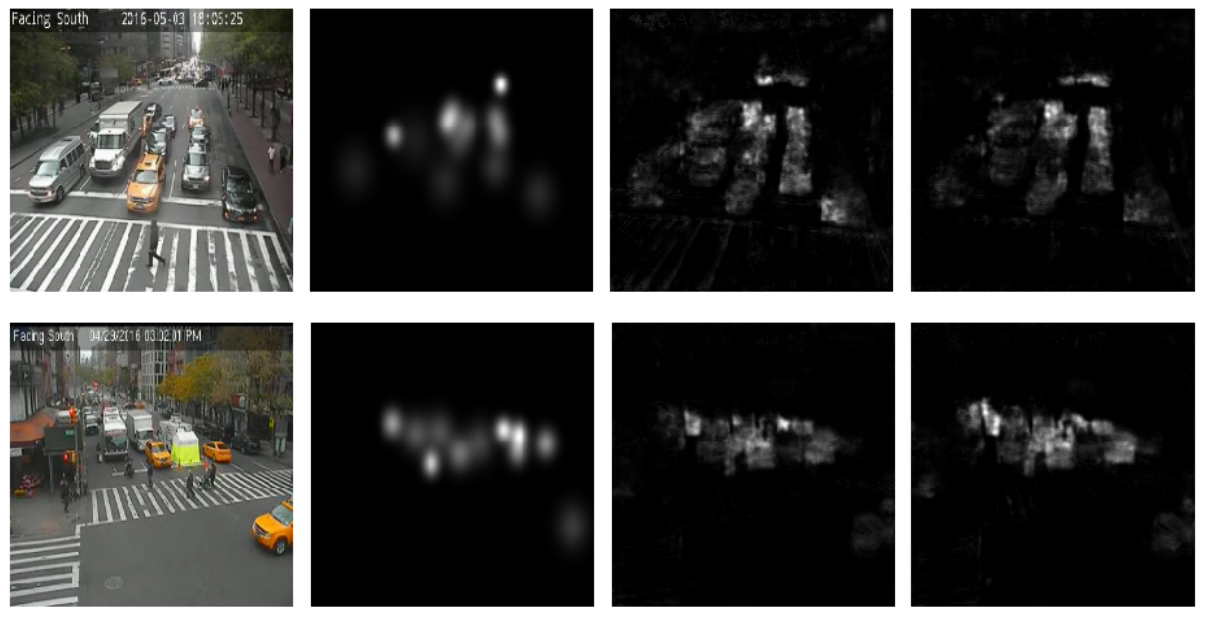}}
\caption{Examples of predicted density maps belonging to two samples of the validation subset (each row corresponds to a sample). From left to right, the original image, the ground truth density map, the predicted density map obtained using the model without the discriminator, and the predicted density map using our domain adaptation algorithm.} 
\label{example_outputs}
\end{figure}

Since all the metrics that we considered take into account only the counting errors, we also plot some examples of the predicted density maps using our model either with and without the discriminator. Figure \ref{example_outputs} shows the ground truth and the predicted density maps for two random samples of the validation subset. As we can see, the density maps predicted using the model with the discriminator show a decrease of the noise compared with the ones obtained using the baseline model without the discriminator. 

\section{CONCLUSIONS}
In this article, we tackle the problem of determining the density and the number of objects present in large sets of images. Building on a CNN-based density estimator, the proposed methodology can generalize to new sources of data for which there is no training data available. We achieve this generalization by adversarial learning, whereby a discriminator attached to the output induces similar density distribution in the target and source domains. Experiments show a significant improvement relative to the performance of the model without domain adaptation. Given the conventional structure of the estimator, the improvement obtained by just monitoring the output entails a great capacity to generalize training, thus suggesting applying similar principles to the inner layers of the network. In our view, this work's surprising outcome opens new perspectives to deal with the scalability of learning methods for large physical systems with scarce supervisory resources.

\ack This work was partially supported by LARSyS - FCT Plurianual funding 2020-2023 and by H2020 project AI4EU under GA 825619.

\bibliography{ecai_long}
\end{document}